# Image Pre-processing on NumtaDB for Bengali Handwritten Digit Recognition

Ovi Paul
Department of Computer Science and Engineering
Independent University, Bangladesh
ovipaulcs@gmail.com

*Abstract*—**NumtaDB is by far the largest data-set collection for handwritten digits in Bengali. This is a diverse dataset containing more than 85000 images. But this diversity also makes this dataset very difficult to work with. The goal of this paper is to find the benchmark for pre-processed images which gives good accuracy on any machine learning models. The reason being, there is not any available pre-processed data for Bengali digit recognition to work with like the English digits for MNIST.**

## I. INTRODUCTION

The aim of this research paper is to find a benchmark for image pre-processing of Bengali digits. By observing the accuracy in different methods on this pre-processed data, the effectiveness of it can be measured. Many researches have been done on Bengali handwritten digit recognition before [7]. These research papers include digital image processing, neural network etc. Datasets that has been used to conduct these researches are CMATERDB and ISI datasets. But none have been done on NumtaDB[6] dataset before.

In this paper, the main content centres around image pre-processing. Finding out the methods that will clean the images from any noise, spots, gridlines etc. Then implementing those methods or techniques to get clean images which will be the final pre-processed data that will be worked on. After finally getting all the data pre-processing work, the final step is to put it through few training models. Only the standard training models will be used, which are often used for image classifications or just classification problems like this one.

## II. LITERATURE REVIEW

Currently over 250 million people speaks Bengali natively. Even to this day the number is increasing rapidly. With increasing number of people speaking Bengali, comes the need for advance technology in this era.

Optical Character Recognition is one such system that allows to scan printed, typewritten or handwritten text (numerals, letters or symbols) and convert scanned image in to a computer process able format, either in the form of a plain text or a word document. OCR is used when creating a similar document in paper as a document in electronic form takes more time. The converted text files take less space than the original image file and can be indexed. Hence the use of OCR adds an advantage to the user who had to deal with conversion of great amount of paper works in to electronic form. Another important goal to achieve is to make a standard dataset for Bengali digits, like the existing MNIST database [1]. By benchmarking it, this preprocessed data can be used in any research works directly. For any applications requiring dataset for digits, the pre-processed data can be directly trained for that work. No extensive research or resource needs to be applied here.

To produce these applications, algorithms are needed to train the device. Even before that pre-processing of those images are needed. After determining the proper pre-processing and training techniques, advanced level applications can be attempted to create from this study.

## III. DATASET

Before NumtaDB, ISI and CMATERDB was the only two datasets for Bengali digit recognition. ISI dataset was the largest dataset for Bengali digits consisting of over 23,000 images. But NumtaDB has set a new milestone by providing with more than 85,000 images of Bengali digits. The MNIST database contains 60,000 training images and 10,000 testing images, setting a total of 70,000 images. Even MNIST, the most popular English digit dataset contains lesser number of images than NumtaDB

TABLE I. ALL DATASETS

| Dataset Name | Dataset Number |
|---|---|
| ISI | 23299 |
| CMATERDB | 6000 |
| NumtaDB | 85596 |

NumtaDB contains 5 datasets for training and 6 datasets for testing the given model. The 5 datasets both for training and testing were provided by Bengali Handwritten Digits Database (BHDDB), BUET101 Database (B101DB), OngkoDB, DUISRT and Bangla Lekha-Isolated. These datasets were marked as 'a', 'b', 'c', 'd' and 'e'. For the testing there was another dataset that was not included in the training dataset. The last 'f' dataset for testing was provided by UIUDB. Only dataset from UIUDB has been used only for testing but for the other sets, those were split 85% to train and 15% to test. A total of 72044 number of images were used for training the models. For testing all the 6 datasets makes a total of 13552 images. For this research topic both the training dataset and testing dataset had to be pre-processed i.e. a total of 85596 images had to be pre-processed.

TABLE II. ENGLISH AND BENGALI NUMERALS

| English Digits | Bengali Digits | Sample from NumtaDB |
|---|---|---|
| 0 | ০ | |
| 1 | ১ | |
| 2 | ২ | |
| 3 | ৩ | |
| 4 | ৪ | |
| 5 | ৫ | |
| 6 | ৬ | |
| 7 | ৭ | |
| 8 | ৮ | |
| 9 | ৯ | |

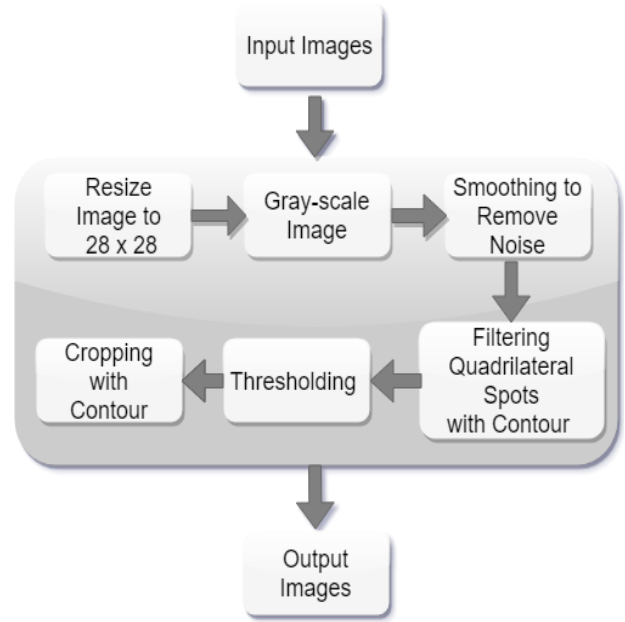

Fig.1.  Image Pre-processing Dataflow

The sample provided to table II, is just a fraction of the whole database. By going through more images in this database, more diverse handwritten digits can be seen. Working with huge data is very advantageous for doing research work. But there are many challenges to tackle to work with NumtaDB dataset. As many images are heavily augmented and contains noise, it is very much difficult to produce good accuracy with this dataset. By benchmarking a pre-processing technique, later works i.e. processing and model training will be much easier to accomplish and understand.

## IV. PROPROSED APPROACH FOR IMAGE PRE-PROCESSING

For pre-processing the images, a set of methods were being used. Pre-processing dataflow is shown on the fig.1. Process for all 5-training dataset and 6-testing dataset has the same set of pre-processing technique. For the whole dataset the dataflow of images needs to be one after the another according to the fig.1. If the sequence of the dataflow is changed or interrupted, then the images will remain distorted. After extensive research on the topic, these set of methods proved to be most useful.

Before all other methods resizing is done at the very start to reduce the computational cost of later processes. Doing this method later would increase the computational cost significantly. Then all images are converted to grayscale to keep all the images similar types, as some images in the dataset were coloured and some were already grayscale. Then median blurring is used to remove or blur the noise on the background. Then by contouring quadrilateral shapes, the spots are removed from the images that contained quadrilateral black coloured spots. Again, thresholding is done to keep the background and foreground colour of all digits in the image same. At last, the digits are contoured this time to focus on it and keep it on the centre and crop the digit only, removing all the redundant data. Finally, the desired output image is received.

### A. RESIZE IMAGE

In NumtaDB there are 5 set of training images and 6 sets of testing images. All the sets contain images with different resolution. As it is not possible to train data with different resolution, all images needed to be resized.

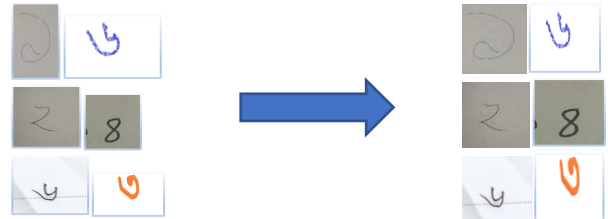

Fig.2.  Resizing Image to 28 x 28

Images with higher resolution consumes more time and resources, so a low resolution will be fixed for all images. For pre-processing this database, the chosen resolution of the images will be 28 x 28 pixels. Both the width and height of all the images will be 28 pixels. That rounds to 784 pixels in total. By resizing the images, the computational cost will be greatly reduced, and the uniformity will be maintained.

### B. GRAYSCALE IMAGE

The database contains images with various colours. The digits in the images can be found in colour black, white, blue, green, orange etc. This creates a problem, as same digits with different colours can be misclassified easily. Since different colours have different value of RGB. For that reason, all the images will be converted to grayscale.

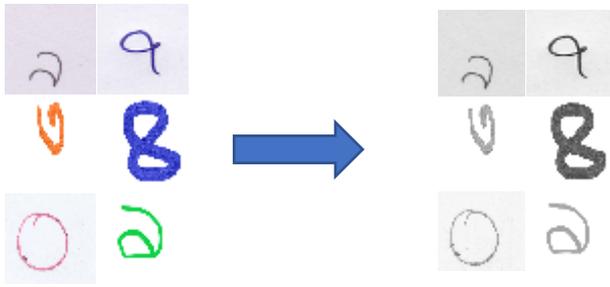

Fig.3. Converting All Image to Grayscale

As it is easier to calculate the values between 0 to 255 than calculating values of all three channels for RGB.

### C. SMOOTHING FOR NOISE REMOVAL

Many of the images contained lot of noises. To remove those noise, Median Blurring [] was used.

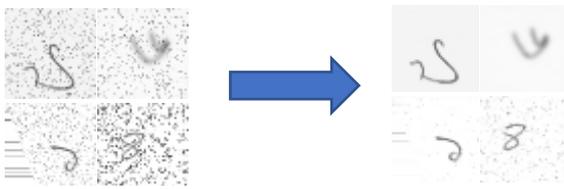

Fig.4. Applying Median Blur on Images

Reason for using Median Blur, it is most effective on salt and pepper noise which contains on these images.

### D. FILTERING QUADRILATERAL SPOTS WITH CONTOUR

Some of the images are heavily distorted with black spots. There are few types of shapes that were found on these images e.g. trapezoid, rectangle, parallelogram, square etc. But all of these shapes contain four sides, so all of them are quadrilateral. For few images there were spots containing more than four sides. But those images look like overlapped quadrilaterals, one top of the another one. So, the applied algorithm to detect quadrilaterals works on the overlapped quadrilaterals generating more than four sides.

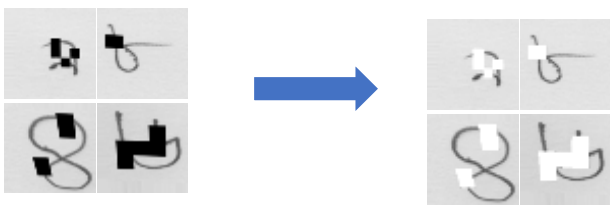

Fig.5. Removing Spots

After detecting the desired contours [8], the detected area is converted to white. Reason for converting the spot to white is to differentiate against the digit colours. Also, the background being white, the spot will blend in with the background.

### E. THRESHOLDING WITH CONTOUR

As mentioned before, background and foreground colour of the images need to be same, for the training models to understand. Even after converting images to grayscale, there were still some problems left. Images shown left side the fig.6., are all grayscale. But for some of the images foreground is black and background is white and vice-versa. So, grayscale values of background and foreground becomes different even though all are grayscale. To avoid this problematic situation, implementation of Otsu's Thresholding [11] is needed. The concept of this thresholding has come from the Otsu's algorithm [10]. A threshold value is passed, which creates a barrier between two peaks of values. One peak being on 255 and another on 0. For this database, the threshold value passed was 127. Then all the pixels in an image was divided by counting pixel values less than 127 and more than or equal 127. The majority side of the pixels would then be converted to value 0, which will be the background and minority to 255 which is the foreground.

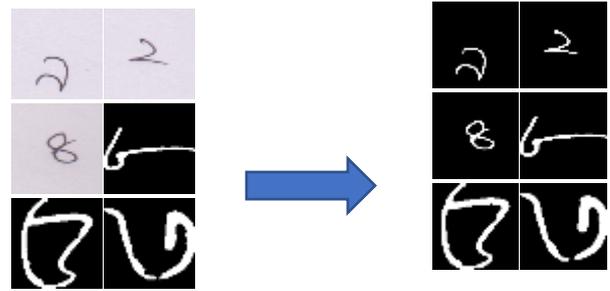

Fig.6. Thresholding to get Same Background and Foreground

### F. CROPPING IMAGE WITH CONTOUR

It is easier for machine learning models to understand or train images with target being on the centre and more zoomed in. NumtaDB contains only images of isolates digits. In those images, there might be shadow of another digit, noise, grids etc. but the largest object is still the targeted isolated digits. So, by detecting the largest contour [9], the targeted digit is being detected. The edges of the detected digit are calculated and data or pixels beyond those edges are cropped out.

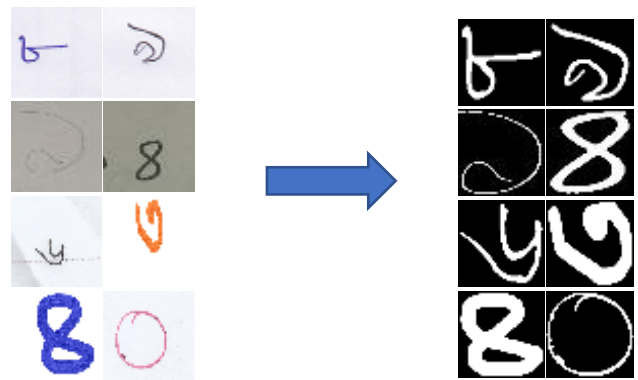

Fig.7. Cropping Digits

The images contain a lot of redundant data that is not needed. Removal of those data can decrease the computational cost significantly.

## V. INSIGHTS FROM VARIOUS MODELS

### A. ACCURACY

By observing accuracy of various machine learning models before and after the pre-processing, a great deal of change can be seen [19]. The accuracy of most of the models increased significantly higher without making any changes or tweaks to those models. Model with the most boost in accuracy came from Support Vector Machine (SVM).

TABLE III.  COMPARING ACCURACY

| Training Model Name | Accuracy | |
|---|---|---|
| | *Non-pre-processed* | *Pre-processed* |
| Convolutional Neural Network (CNN) | 0.68182 | 0.91323 |
| Capsule Network (CapsNet) | 0.65233 | 0.87107 |
| K-Nearest Neighbour (KNN) | 0.49035 | 0.77967 |
| K-Nearest Neighbour with Principle Component Analysis (KNN with PCA) | 0.41836 | 0.79812 |
| Support vector machine (SVM) | 0.20845 | 0.84538 |
| Support vector machine with Principle Component Analysis (SVM with PCA) | 0.52507 | 0.74254 |
| Logistic Regression | 0.31933 | 0.74254 |
| Logistic Regression with Principle Component Analysis | 0.23458 | 0.73854 |
| Decision Tree | 0.3184 | 0.70476 |

### B. COMPUTATIONAL TIME

Along with the accuracy, the computational cost has been reduced as well [5]. For some models it is a minor change but for other models the changes are major.

TABLE IV.  COMPARING COMPUTATIONAL COST

| Training Model Name | Computational Time Cost | |
|---|---|---|
| | *Non-pre-processed* | *Pre-processed* |
| Convolutional Neural Network (CNN) | 378.82 minutes | 293.12 minutes |
| Capsule Network (Caps Net) | 412.22 minutes | 357.32 minutes |
| K-Nearest Neighbour (KNN) | 19.31 minutes | 16.55 minutes |
| K-Nearest Neighbour with Principle Component Analysis (KNN with PCA) | 10.51 minutes | 8.53 minutes |
| Support vector machine (SVM) | 304.55 minutes | 107.42 minutes |
| Support vector machine with Principle Component Analysis (SVM with PCA) | 16.53 minutes | 3.20 minutes |
| Logistic Regression | 1.10 minutes | 0.44 minutes |
| Logistic Regression with Principle Component Analysis | 35.16 minutes | 32.42 minutes |
| Decision Tree | 7.47 minutes | 1.58 minutes |

Even though the non-pre-processed data is resized and applied grayscale on it to run on all the models, still it takes more time comparing to pre-processed data.

## VI. RESULT AND ANALYSIS

By following this methodology for image pre-processing, the final data received can easily score over 70% accuracy. The methodology or approach can be counted as success, as the 70% accuracy coming from machine learning models that gives bad accuracy on heavily noised and augmented images. For this reason, without pre-processing images, the accuracy on most of the training models shows low on accuracy. The data without pre-processed gives accuracy starting from 20%. The data in which pre-processing is not done is certainly generates more misclassified results, as the maximum accuracy was just 68%. For the best score or accuracy from pre-processed data, is the Convolutional Neural Network (CNN) [12].

In terms of accuracy the neural network models CNN and Capsule Network (CapsNet) had a good score for both non-pre-processed data and pre-processed data. These neural network models are used often more image classification problems as these models give better accuracy than other machine learning models. Neural network models can be tweaked in many ways, which is the reason for its high accuracy. The number of layers, hyper parameter, training time etc. can be tweaked to get higher accuracy. But as the aim of this research was to observe the accuracy from standard models, no tweak was made to these neural network models. A standard one input, one hidden and one output layer was used and without any tweaks in the parameter. Still the models managed to cross the 90% mark. Purpose of this research was to find an image pre-processing benchmark for NumtaDB. The set of techniques used for this database has certainly showed good results to benchmark it.

## VII. DISCUSSION

From Table III and Table IV, the difference between non-pre-processed data and pre-processed data can be easily understood. The change in accuracy is something previously discussed. For all the machine learning models, the increase in accuracy was significant. For models like Support Vector Machine (SVM) [18] the accuracy went to 84% from the previous 20%. Approximately 60% increase in accuracy occurred for this specific model.

Now, for the computational cost, pre-processed images took less time comparing to non-pre-processed images. Again, for machine learning models like SVM [17], the computational cost reduced from approximately 304 minutes to approximately 107 minutes. This is a significant amount of change. For other machine learning models, there were difference in computation time too. The reason for such reduction in computational time is for removing the redundant data or pixels. During the pre-processing of NumtaDB, images with redundant data or pixels were cropped out. Because approximately all the images were cropped, it was easier to train on all those images for all the machine learning models.

## VIII. FUTURE WORK

Previously, removal of spotted images was mentioned. For most images, lose of those pixels gave a boost in the accuracy, as all those pixels contained noise data. But for few images the same concept is not true.

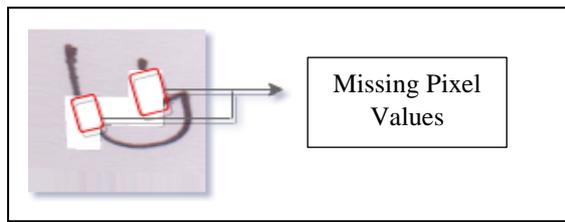

Fig.8. Losing Data

Figure.8. shows that for the Bengali digit '6', some important data are being missing. The spot for this image overlapped on a large portion of the digit. By removing the spot, a portion of the digit is being removed. Images with large portion of the pixel data missing for digits will get a higher chance to be misclassified by the machine learning models. So, this kind of spotted images will have high chance of getting misclassified, whether the spot is removed or not. Also, some testing images were augmented to make those images rotate. It is difficult for training models to correctly classify those. For now, the augmented data generator is a solution to augment your original training images and them train on it. But to recognize or classify these images from just pre-processed data, better solution to this problem needs to be developed. For these reason, a set of better pre-processing technique needs to be applied on these types of images. By doing so, the accuracy just from pre-processing the images will increase more.